\documentclass[journal]{IEEEtran}

\usepackage{multirow}
\usepackage{amsmath}
\usepackage{amssymb}
\usepackage{cite}
\usepackage{amsfonts}
\ifCLASSINFOpdf
   \usepackage[pdftex]{graphicx}
   \graphicspath{{../pdf/}{../jpeg/}}
   \DeclareGraphicsExtensions{.pdf,.jpeg,.png}
\else
   \usepackage[dvips]{graphicx}
   \graphicspath{{../eps/}}
   \DeclareGraphicsExtensions{.eps}
\fi

\usepackage{amsmath}

\usepackage{algorithmic}
\usepackage{array}
\ifCLASSOPTIONcompsoc
 \usepackage[caption=false,font=normalsize,labelfont=sf,textfont=sf]{subfig}
\else
 \usepackage[caption=false,font=footnotesize]{subfig}
\fi

\usepackage{fixltx2e}

\newcommand{\ie}[1]{\textit{i.e.}, #1}
\newcommand{\eg}[1]{\textit{e.g.}, #1}

\usepackage{xcolor}

\begin{document}
\title{A Hybrid Brain-Computer Interface Using Motor Imagery and SSVEP Based on Convolutional Neural Network}

\author{Wenwei Luo,
        Wanguang Yin,
        Quanying Liu$^{\ast}$,
        Youzhi Qu$^{\ast}$% <-this % stops a space
\thanks{This work was funded in part by the National Natural Science Foundation of China (62001205), National Key R\&D Program of China (2021YFF1200804), Shenzhen Science and Technology Innovation Committee (20200925155957004, KCXFZ2020122117340001), Shenzhen-Hong Kong-Macao Science and Technology Innovation Project (SGDX2020110309280100), Shenzhen Key Laboratory of Smart Healthcare Engineering (ZDSYS20200811144003009), Guangdong Provincial Key Laboratory of Advanced Biomaterials (2022B1212010003).}% <-this % stops a space
\thanks{W. Luo, W. Yin, Q. Liu and Y. Qu are with Shenzhen Key Laboratory of Smart Healthcare Engineering, Guangdong Provincial Key Laboratory of Advanced Biomaterials, Department of Biomedical Engineering, Southern University of Science and Technology. 
% W. Yin is with **** \todo{check}
}
\thanks{$^{\ast}$ Corresponding authors: Y. Qu ({\tt\small quyz@mail.sustech.edu.cn})  \& Q. Liu ({\tt\small liuqy@sustech.edu.cn}) }
}


\markboth{Luo, Yin, Liu \& Qu (2022)}%
{Luo \MakeLowercase{\textit{et al.}}: CNN-based Hybrid BCI using MI and SSVEP}



\maketitle

\begin{abstract}
The key to electroencephalography (EEG)-based brain-computer interface (BCI) lies in neural decoding, and its accuracy can be improved by using hybrid BCI paradigms, that is, fusing multiple paradigms.
However, hybrid BCIs usually require separate processing processes for EEG signals in each paradigm, which greatly reduces the efficiency of EEG feature extraction and the generalizability of the model.
Here, we propose a two-stream convolutional neural network (TSCNN) based hybrid brain-computer interface. It combines steady-state visual evoked potential (SSVEP) and motor imagery (MI) paradigms. TSCNN automatically learns to extract EEG features in the two paradigms in the training process, and improves the decoding accuracy by 25.4\% compared with the MI mode, and 2.6\% compared with SSVEP mode in the test data. Moreover, the versatility of TSCNN is verified as it provides considerable performance in both single-mode (70.2\% for MI, 93.0\% for SSVEP) and hybrid-mode scenarios (95.6\% for MI-SSVEP hybrid). Our work will facilitate the real-world applications of EEG-based BCI systems. 
\end{abstract}


\IEEEpeerreviewmaketitle

\section{Introduction}

\IEEEPARstart{B}{rain-computer} interface (BCI) decodes the brain signals (\eg EEG, fMRI, fNIRS) for establishing direct communication between the human brain and computer (or more generally, between brain and machine) to replace traditional brain signal output pathways, such as peripheral nerve and muscle tissues. In particular, EEG-based BCI is to detect EEG features related to mental states or intentions and translates these features into specific commands. A number of existing works have focused on improving EEG decoding performance in terms of accuracy~\cite{yin2022partial}, robustness~\cite{robust}, efficiency~\cite{efficiency}, and cross-subject reliability~\cite{subj}.

Typical EEG-based BCI paradigms include the motor imagery (MI)~\cite{mi1, mi2, yin2022partial}, P300~\cite{p3001, p3002}, and steady-state visual evoked potential (SSVEP)~\cite{ssvep1, ssvep2, yin2022partial}.
During motor imagery, such as imagining limb movements without actual limb movements, the motor cortex is activated, producing neural patterns similar to those produced during actual motor execution~\cite{mi_mec1, mi_mec2}. These neural patterns can be captured with EEG recordings.
Traditional human-computer interaction methods, such as using a keyboard to interact with a computer, require the user to perform actual operations to control.
MI-based BCI provides a new mode of human-computer interaction for virtual reality and the manipulation of external devices without actual action execution. MI-based BCI also can control rehabilitation equipment such as wheelchairs and prosthetic devices for people with motor disabilities~\cite{mi_wheel, mi_prosthetic}.
SSVEP signals are the entrained neural response to visual stimuli at specific frequencies, which can be used for neural decoding.
As a widely-used BCI paradigm, SSVEP has the advantages of fast recognition speed, high accuracy, and extendable classes of commands for external device control and text spelling~\cite{ssvep_spell}.
However, almost all of these BCI paradigms have their own limitations. MI-based BCI suffers from its unsatisfactory decoding accuracy and few recognizable command classes. SSVEP-based BCI requires the subjects to stare at the monitor for a long time, and the flickering stimuli easily cause visual fatigue~\cite{bci_vr}. Therefore, a more reliable, efficient and comfortable BCI paradigm is urgently needed.

\begin{figure*}[!t]
    \centering
    \includegraphics[width=\linewidth]{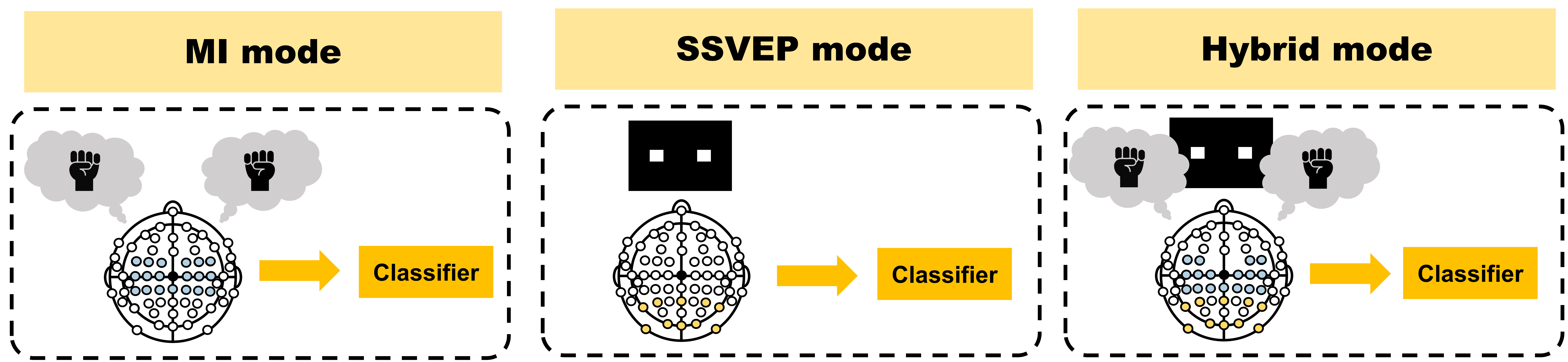}
    \caption{Three modes of BCI system. (left) MI mode, in which the subject imagines the left- or right-hand movement. (middle) SSVEP mode, in which the subject stares at the left or right flickering stimulus. (right) Hybrid mode, in which the subject stares at the left or right flickering stimulus while imagining the corresponding hand movement.}
    \label{mode}
\end{figure*}

Recently, fusing multiple BCI paradigms (\eg MI and SSVEP) and fusing multimodal neural signals have been proposed, called \textit{hybrid BCI}.
Some studies have recently reported using hybrid BCI can improve BCI performance, achieving better efficiency and accuracy compared to single mode BCI~\cite{yin2022partial, eeg-fnir, eeg-fmri}. 
Moreover, inter-subject reliability is a major consideration in BCI real-world applications. The subjects not suitable for a certain BCI paradigm may fit well in other paradigms. For instance, Allison et al. \cite{Allison_2010} has reported that some subjects who fail to produce recognizable brain activity patterns in the MI-based BCI paradigm can perform well in SSVEP-based BCI, and vice versa. 
Subjects even if unable to produce recognizable neural patterns in any single mode (either MI or SSVEP) may have recognizable neural patterns in hybrid mode.

Inspired by the recent advances in deep learning and their universality in computer vision and natural language processing, deep learning models have attracted more and more attention in neural signal processing and neural decoding~\cite{yu2022embedding, qu2022transfer}.
Deep learning to extract EEG features can naturally learn to combine and utilize complex information in SSVEP and MI modes, providing a general framework for multi-paradigm BCI. 
Traditional neural decoding methods in BCI systems, such as canonical correlation analysis (CCA), common spatial patterns (CSP), and filter bank CSP (FBCSP), need to design different algorithms to process EEG signals in specific paradigms, and most traditional methods rely on prior information and are susceptible to noise interference.
In contrast, deep learning-based neural decoding can effectively extract EEG features in the time domain and frequency domain without the necessity of prior knowledge of EEG feature engineering. 
More importantly, deep learning can utilize distributed and hierarchical features through multiple layers of nonlinear information processing, resulting in higher performance than traditional methods.
For instance, Kwon et al.\cite{subj} has proposed a subject-independent convolutional neural network (CNN) framework for an MI-based BCI system, achieving much higher accuracy compared to traditional methods.
Schirrmeister et al.\cite{Sch} has implemented a range of CNN architectures for decoding MI from raw EEG signals and visualized the extracted EEG features to study the latent information learned by CNNs and how feature representations facilitate neural decoding.

\begin{figure*}[!t]
    \centering
    \includegraphics[width=\linewidth]{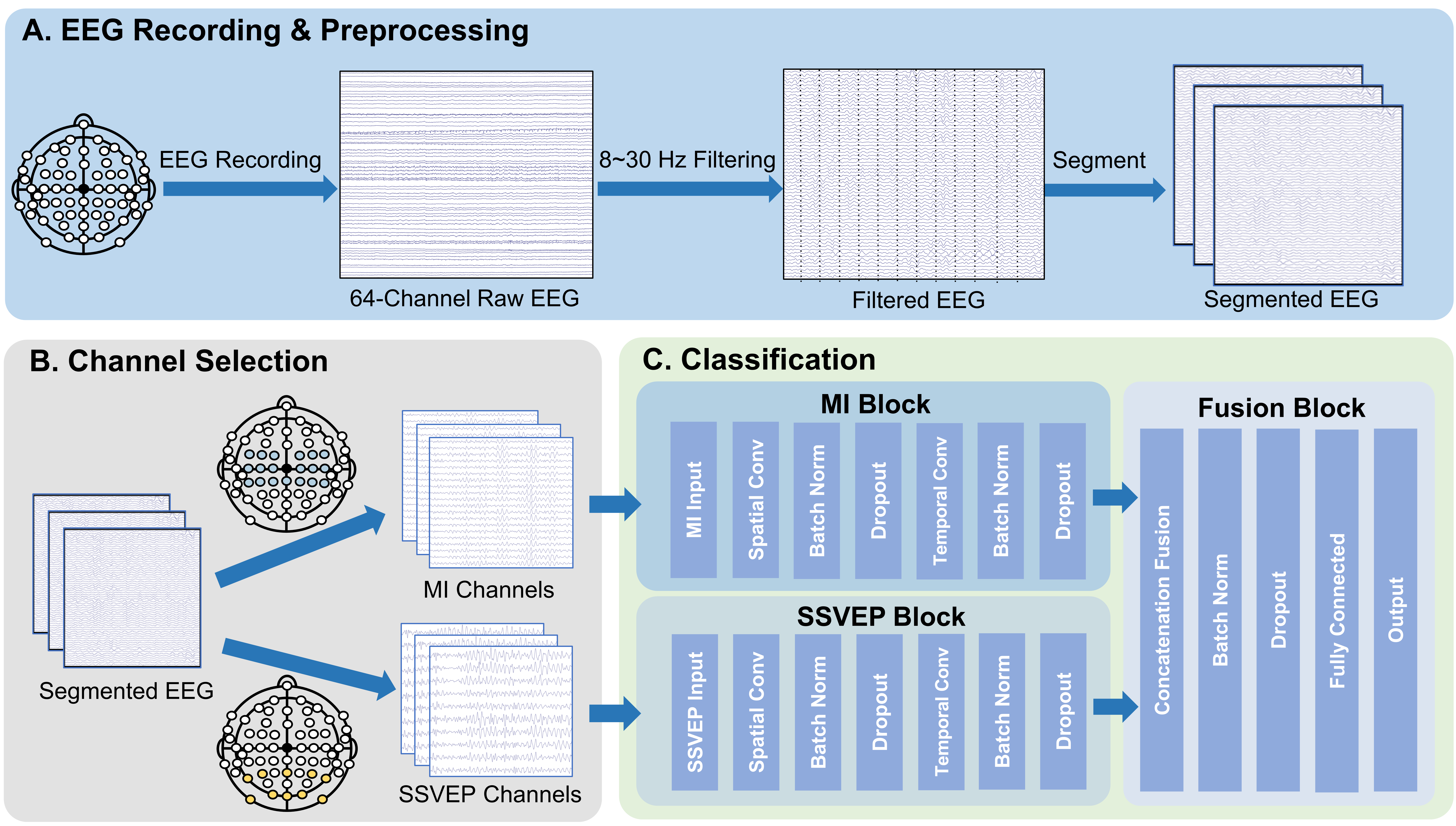}
    \caption{The TSCNN framework: \textbf{A}. EEG recording and preprocessing: the recorded EEG is filtered by $8\sim 30 Hz$ bandpass filter, and then segmented into 4s length.  \textbf{B}. Channel selection: select the specific channels of MI and SSVEP. \textbf{C}. Classification: classify the input EEG segments with TSCNN.}
    \label{flow}
\end{figure*}

In this study, we propose a hybrid BCI system based on MI and SSVEP, and a two-stream convolutional neural network (TSCNN) which is used to decode EEG. As shown in Fig. \ref{mode}, TSCNN can be used in three modes (\ie MI, SSVEP, and hybrid). The main contributions of this study can be summarized as follows.
\begin{itemize}
    \item TSCNN can automatically learn to extract the EEG features in the two BCI paradigms during the training process, avoiding the limitation of the traditional methods to design specific algorithms to extract features in different paradigms.
    % \item TSCNN is more versatile, as it can be used in single-mode BCI paradigms and hybrid BCI.
    \item We present a novel training strategy that enables TSCNN to achieve high decoding accuracy in hybrid mode while maximally reserving performance in MI mode and SSVEP mode.
    \item Through the training strategy and the two-stream architecture, TSCNN is more versatile as it provides considerable performance in both single-mode and hybrid-mode scenarios.
\end{itemize}

\section{Materials and Methods}

\subsection{Dataset Description}
The EEG dataset used in our study was collected by the Department of Brain and Cognitive Engineering at Korea University. It includes data from 54 subjects performing a binary-class MI task and a four-class SSVEP task. The dataset consists of two sessions, each of which includes an offline training phase and an online testing phase. For this study, we use the training phase of session$_1$.
The EEG signals were recorded at a sampling rate of 1,000 Hz using 62 Ag/AgCl electrodes and a BrainAmp amplifier (Brain Products; Munich, Germany).

In the MI task, a trial starts with a black fixation cross displayed in the center of the screen, which lasts for 3 seconds, followed by a left or right arrow as a visual cue, lasting for 4 seconds during which the subject imagine a grasping movement with the corresponding hand. In the end of each trial, the screen remains blank for 6s ($\pm$1.5s). The MI paradigm consists of 100 trials, including 50 left MI trials and 50 right MI trials. In the SSVEP task, four target SSVEP stimuli flickering at 5.45Hz, 6.67Hz, 8.57Hz, and 12Hz are presented at four locations (down, right, left, and up, respectively) on the monitor. Participants are instructed to fixate at the center of the black screen and then gaze in the direction where the target stimulus is highlighted in different colors. In each trial, SSVEP stimulus is presented for 4s followed by a 2s rest. The SSVEP paradigm consists of 100 trials, including 25 trials of each frequency.

The preprocessing and channel selection of EEG data are shown in Fig. \ref{flow}. In this study, we aim to combine MI and SSVEP to achieve left-right binary classification, so the frequencies representing left-right classes ($6.67\mathrm{Hz}$ and $8.57\mathrm{Hz}$) in the SSVEP task are used. In addition, 20 electrodes in the motor cortex region (FC-5/3/1/2/4/6, C-5/3/1/z/2/4/6, and CP-5/3/1/z/2/4/6) and 10 electrodes in the occipital region (P-7/3/z/4/8, PO-9/10, and O-1/z/2) are selected for MI and SSVEP tasks, respectively.
The MI signals are band-pass filtered between $8 \sim 30$ Hz with a 5th-order Butterworth digital filter. Both MI and SSVEP signals are segmented from $0 \sim 4,000$ms.

\subsection{Model Architecture}
TSCNN consists of three main blocks: the MI block, SSVEP block, and Fusion block. Fig. \ref{flow}C illustrates the TSCNN architecture in this study. The MI block and the SSVEP block share a similar architecture adopted from the single-stream CNN (SCNN)~\cite{ssvep1}. 
 
The MI block and the SSVEP block have two main layers: a spatial convolutional layer and a temporal convolutional layer.
EEG channels are selected for MI and SSVEP tasks specifically, and these channels are input into the corresponding blocks, as shown in Fig.~\ref{flow}B.
The input dimension is $N_{ch}\times N_{t}$, where $N_{ch}$ is the number of channels and $N_{t}$ is the number of time components.
The spatial convolutional layer that performs 1D convolutions across the channel dimension with the kernel dimension of $20 \times 1$ and $10 \times 1$ corresponding to MI and SSVEP.
The temporal convolutional layer operates on the spectral representation of the input with a kernel dimension of $1\times 10$. 
After the spatial and temporal convolution, the feature map dimension of MI and SSVEP blocks are both $1\times (N_{t}-10+1)$. 
Then the feature maps of MI and SSVEP blocks are concatenated in the fusion block. 
Afterward, the concatenated feature maps are connected to a fully connected layer for classification, followed by batch normalization and dropout. Specifically, batch normalization is used to accelerate training and improve the generation performance by reducing the internal covariance shift in the input, turning it into a standard normal distribution $\sim N(0, 1)$\cite{bn}. Dropout is used to avoid overfitting and improve generalizability~\cite{drop1, drop2, dropout}.
Finally, the decision of TSCNN is made by a sigmoid function.

\begin{table*}[t]
\centering
\caption{Decoding performance (mean±std) of three models}
\setlength{\tabcolsep}{6mm}
\begin{tabular}{cccccc}
\hline
Model & Paradigm & Accuracy & Sensitivity & Specificity & MSE\\ 
\hline
\multirow{2}{*}{SCNN} & $MI$ & 0.704 $\pm$ 0.008 & 0.703 $\pm$ 0.020 & 0.706 $\pm$ 0.023 & 0.575 $\pm$ 0.003\\
& $SSVEP$ & 0.937 $\pm$ 0.003 & 0.931 $\pm$ 0.010 & 0.944 $\pm$ 0.009 & 0.119 $\pm$ 0.00\\
\hline
 & $MI$ & 0.678 $\pm$ 0.017 & 0.764 $\pm$ 0.078 & 0.591 $\pm$ 0.099 & 0.609 $\pm$ 0.019\\
TSCNN$_1$ & $SSVEP$ & 0.934 $\pm$ 0.004 & 0.937 $\pm$ 0.018 & 0.931 $\pm$ 0.01 & 0.126 $\pm$ 0.004\\
 & $Hybrid$ & 0.956 $\pm$ 0.006 & 0.961 $\pm$ 0.011 & \textbf{0.952 $\pm$ 0.018} & \textbf{0.100 $\pm$ 0.005}\\
 \hline
 & $MI$ & 0.702 $\pm$ 0.007 & 0.709 $\pm$ 0.028 & 0.695 $\pm$ 0.024 & 0.583 $\pm$ 0.006\\
TSCNN$_2$ & $SSVEP$ & 0.930 $\pm$ 0.006 & 0.943 $\pm$ 0.013 & 0.916 $\pm$ 0.020 & 0.134 $\pm$ 0.005\\
 & $Hybrid$ & \textbf{0.956 $\pm$ 0.006} & \textbf{0.964 $\pm$ 0.007} & 0.948 $\pm$ 0.013 & 0.110 $\pm$ 0.004\\
\hline
\end{tabular}
\label{met}
\end{table*}

\subsection{Feature Representation in TSCNN}
We denote $\textbf{X}=\{ \textbf{\textit{x}}_i\}_{i=1}^n \in \mathbb{R}^{N_{ch}\times N_{t}}$ as a set of preprocessed single trial EEG, and let $\textbf{Y}=\{ \textbf{y}_i\}_{i=1}^n$ be the matching class labels, where n is the number of trials, $\textbf{x}_i$ is the single-trial EEG, and $\textbf{\textit{y}}_i$ is the class label of the single trial.
The selected MI channels and SSVEP channels from 64-channel EEGs of a subject $\textbf{X}$ are selected (Fig. \ref{flow}B), denoted as $\textbf{X}_M$ and $\textbf{X}_S$, respectively. We denote $\textbf{X}_M$ and $\textbf{X}_S$ as the inputs of the MI block and SSVEP block, respectively.
Given inputs $\textbf{X}_M$ and $\textbf{X}_S$, the output of TSCNN is denoted as $\textbf{Y}=f_{T}(\textbf{X}_M, \textbf{X}_S)$, where $f_{T}$ refers to the TSCNN network.

\subsection{Division of Data}
Data from forty subjects are used to train the TSCNN model. We propose two training strategies: using only the hybrid-mode EEG (TSCNN$_1$), and using both the single-mode EEG and hybrid-mode EEG (TSCNN$_2$). 

The first strategy is to train the model using only hybrid-mode EEG. We denote the TSCNN trained in this strategy as TSCNN$_1$. The inputs are presented as:
%One was training in hybrid mode, and the inputs could be presented as:
$$
\textbf{X}_M=
\begin{bmatrix}
\textbf{\textit{x}}_{M1} & \textbf{\textit{x}}_{M2} &\cdots & \textbf{\textit{x}}_{Mn}
\end{bmatrix}
^T
% \hspace{0.5cm}
$$
$$
\textbf{X}_S=
\begin{bmatrix}
\textbf{\textit{x}}_{S1} & \textbf{\textit{x}}_{S2} & \cdots & \textbf{\textit{x}}_{Sn}
\end{bmatrix}
^T
$$
A total number of 2000 samples (40 subjects $\times$ 50 trials) for each paradigm are utilized as inputs for TSCNN$_1$. 10-fold cross-validation is performed in the training session. 

The second strategy uses a combination of the single-mode and hybrid-mode EEG data to train the model. We denote the TSCNN trained in this strategy as TSCNN$_2$. The inputs are presented as:
% The other is training in both hybrid mode and single-MI mode, and the inputs can be noted as:
$$
\textbf{X}_M=
\begin{bmatrix}
\textbf{\textit{x}}_{M1} & \textbf{\textit{x}}_{M2} & \cdots & \textbf{\textit{x}}_{Mn} & \textbf{\textit{x}}_{Mn+1} & \cdots & \textbf{\textit{x}}_{M2n}
\end{bmatrix}
^T
% \hspace{0.5cm}
$$
$$
\textbf{X}_S=
\begin{bmatrix}
\textbf{\textit{x}}_{S1} & \textbf{\textit{x}}_{S2} & \cdots & \textbf{\textit{x}}_{Sn} & \textbf{0} & \cdots & \textbf{0}
\end{bmatrix}
^T
$$
A total number of 4000 samples (40 subjects $\times$ 100 trials) for each paradigm are utilized as inputs for the TSCNN$_2$. 10-fold cross-validation is performed.
%The training fold includes 3600 samples and the validation fold includes 400 samples.

\subsection{Training Parameters}
The weights of the TSCNNs are initialized with a Gaussian distribution $\sim N(0, 0.01)$. The network is trained using backpropagation to minimize the binary cross-entropy (BCE) loss function:
\begin{equation*}
   BCE(y, \hat{y}) = -(y\log (\hat{y})+(1-y)\log(1-\hat{y})),
\end{equation*}
where $y$ denotes the true label and $\hat{y}$ denotes the predicted label. 
All models are trained using the Adam optimizer with a learning rate of 0.00025~\cite{adam1}.
The dropout rate and the batch size are set to 50\% and 64, respectively. All the experiments are conducted on a laptop with AMD-Ryzen 7-5800H 3.20-GHz and 16-GB memory.

\subsection{Evaluation}
EEG data from the rest 14 subjects are used as the test dataset. 
The performance of TSCNN is quantified with the decoding accuracy, sensitivity, specificity and mean square error (MSE). 
A modified SCNN is used as a competing model, whose great performance in single-paradigm classification has been previously reported ~\cite{ssvep1}. 

\section{Results}
\subsection{Decoding performance}
Table \ref{met} shows the decoding performance of SCNN, TSCNN$_1$, and TSCNN$_2$. 
Obviously, TSCNN$_2$ achieves the best decoding accuracy in the hybrid mode and maintains satisfactory performance in the single mode of MI and SSVEP.
Specifically, the averaged decoding accuracy, sensitivity, specificity and MSE from TSCNN$_2$ are $70.2\% (\pm 0.7)$, $70.9\% (\pm 2.8)$, $69.5\% (\pm 2.4)$ and $0.583 (\pm 0.006)$ in MI mode, $93.0\% (\pm 0.6)$, $94.3\% (\pm 1.3)$, $91.6\% (\pm 2.0)$ and $0.134 (\pm 0.005)$ in SSVEP mode, and $95.6\% (\pm 0.6)$, $96.4\% (\pm 0.7)$, $94.8\% (\pm 1.3)$ and $0.11 (\pm 0.004)$ in hybrid mode. 

\begin{figure*}[!t]
    \centering
    \includegraphics[width=\linewidth]{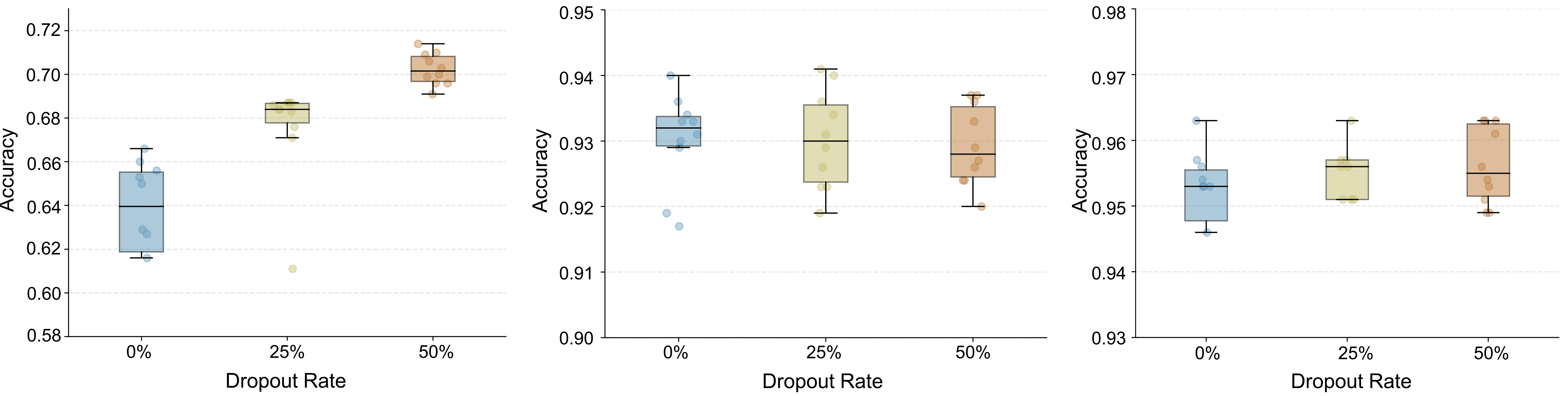}
    \caption{Impact of TSCNN architecture design choices on decoding accuracy in MI mode (left), SSVEP mode (middle), and hybrid mode (right) with different dropout rates. The horizontal axis is the different dropout rates, and the vertical axis is the decoding accuracy. }
    \label{dropout}
\end{figure*}

\begin{figure*}[!t]
    \centering
    \includegraphics[width=\linewidth]{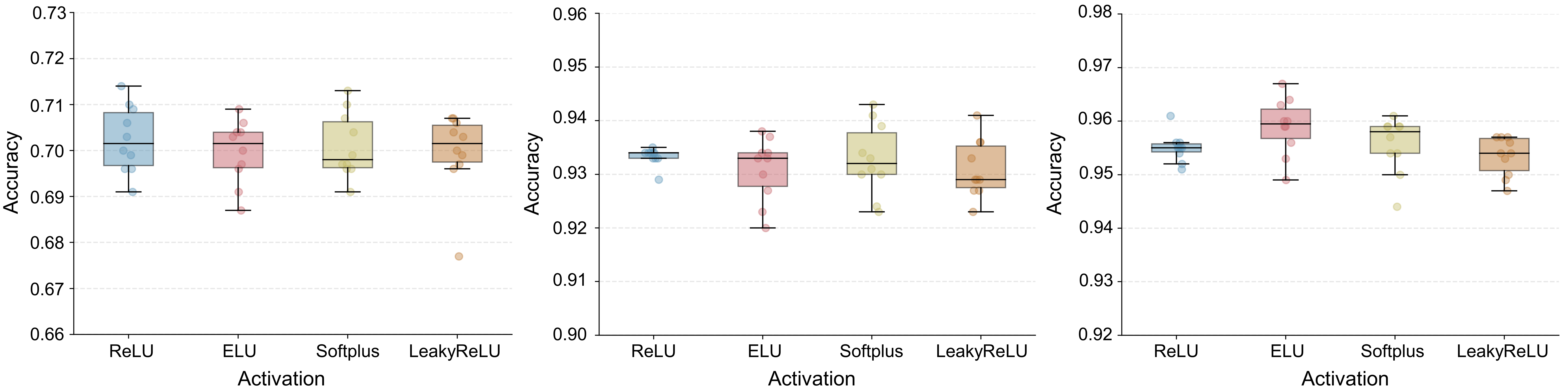}
    \caption{Impact of TSCNN architecture design choices on decoding accuracy in MI mode (left), SSVEP mode (middle), and hybrid mode (right) with different activations. The horizontal axis is the different dropout rates, and the vertical axis is the decoding accuracy.}
    \label{act}
\end{figure*}

Table \ref{fc} shows the performance changes of TSCNN$_2$ in three modes according to the dimension of the fully-connected layer. For MI mode, the decoding accuracies with the change of dimension of the fully-connected layer are $65.1\%$ and $70.4\%$ for no fully connected layer and 16-dimension fully connected layer. For SSVEP mode, they are $93.6\%$ and $93.7\%$ respectively. For hybrid mode, they have the same performance on decoding accuracy, which is $95.6\%$. The model with 16-dimension fully-connected layer performed better overall.

\begin{table}[h]
\centering
\caption{Decoding accuracy of TSCNN$_2$ according to the fully connected number}
\begin{tabular}{|c|c|c|c|}
\hline
Mode & $MI$ & $SSVEP$ & $Hybrid$ \\
\hline
without fully-connected layer & 0.651 & 0.936 & 0.956\\ 
\hline
16-dimension fully-connected layer & \textbf{0.704} & \textbf{0.937} & \textbf{0.956} \\
\hline
\end{tabular}
\label{fc}
\end{table}

Table \ref{fm} illustrates the performance changes of TSCNN$_2$ under three modes according to the number of convolution kernels.
In order to study the effect of the number of convolution kernels, the performance of the TSCNN under different numbers of convolution kernels was tested.
Convolution kernels of $(i, j)$ represent that the spatial convolutional layer and temporal convolutional layer in MI and SSVEP block have $i$ and $j$ convolutional kernels, respectively.
The MI block and SSVEP block have the same structure, and the different numbers of kernels for spatial and temporal convolution are compared to evaluate the performance.
The decoding accuracies for MI mode are $70.4\%$, $66.3\%$, $66.2\%$, $64.6\%$ when the kernel number is (1, 1), (1, 2), (2, 2), (8, 8) respectively.
For SSVEP mode, the decoding accuracies are $93.7\% $, $93.5\% $, $93.3\%$, $93.0\%$ when the kernel number is (1, 1), (1, 2), (2, 2), (8, 8) respectively.
For hybrid mode, the decoding accuracies are $95.6\% $, $95.7\% $, $95.4\%$, $94.7\%$ when the kernel number is (1, 1), (1, 2), (2, 2), (8, 8) respectively. The model with kernel number (1, 1) had the best performance overall.

\begin{table}[!h]
\centering
\caption{Decoding accuracy of TSCNN$_2$ according to the number of convolution kernels}
\begin{tabular}{|c|c|c|c|c|}
\hline
Convolution kernels & (1, 1) & (1, 2) & (2, 2) & (8, 8)\\
\hline
$MI$ & \textbf{0.704} & 0.663 & 0.662 & 0.646\\
\hline
$SSVEP$ & \textbf{0.937} &	0.935 &	0.933 &	0.930\\
\hline
$Hybrid$ & 0.956 & \textbf{0.957} & 0.954 & 0.947\\
\hline
\end{tabular}
\label{fm}
\end{table}

Fig. \ref{dropout} shows the decoding accuracy of TSCNN$_1$ in three modes with 0\%, 25\%, and 50\% dropout rate respectively. From the results, the 50\% dropout rate has a significant improvement in decoding accuracy for the MI mode, while there is no obvious difference in decoding accuracies between different dropout rates in SSVEP and MI modes.

The decoding accuracy of TSCNN$_1$ using different non-linear activation functions are compared. Fig. \ref{act} shows the decoding accuracies of TSCNN$_1$ in the three modes with ReLU~\cite{relu}, ELU\cite{elu}, Softplus\cite{relu}, and LeakyReLU\cite{leakyrelu} activation functions. There is no significant difference in the performance of different activation functions in the three modes.

\begin{figure*}[!t]
    \centering
    \includegraphics[width=0.9\linewidth]{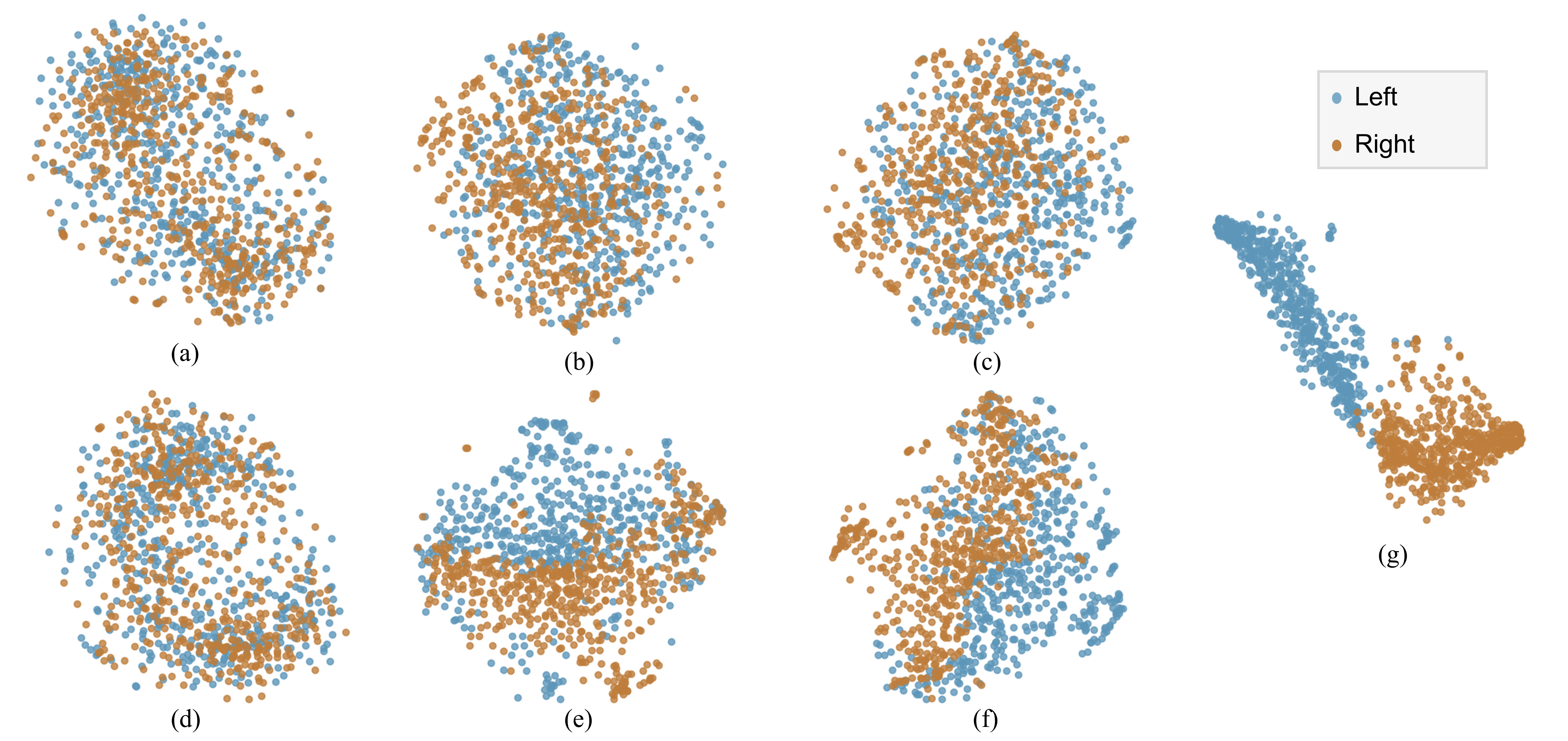}
    \caption{Visualization of features for subjects 20-30 in 2 dimensions using t-SNE$--$TSCNN$_2$. (a)-(c) are the visualization results in MI block. (a) MI input features. (b) Features of spatial convolutional layer in MI block. (c) Features of the temporal convolutional layer in MI block. (d)-(e) are the visualization results in SSVEP block. (d) SSVEP input features. (e) Features of the spatial convolutional layer in SSVEP block.  (f) Features of the temporal convolutional layer in SSVEP block.  (g) is the visualization result of fully-connected layer.}
    \label{tsne}
\end{figure*}

Fig. \ref{tsne} visualized the features at different layers of TSCNN$_2$ in hybrid mode by t-SNE. Fig. \ref{tsne} (a)-(c) are visualization results in the MI block, Fig. \ref{tsne} (d)-(f) are visualization results in the SSVEP block, and Fig. \ref{tsne} (g) is the visualization result in the fusion block. To achieve this, the data of subject 20$\sim$40 is input into pre-trained TSCNN$_2$, and the features were extracted at each layer. Each point in the figure represents a single trial, and the color represents its category (blue represents left and red represents right). The results show that the MI block features are less distinct than the SSVEP block. The features are the most obvious and can be distinguished well after the fusion.

% \section{Discussion}
\subsection{Statistical Analysis} 
We performed a paired sample t-test on the experimental results, and the null hypothesis in this study is that the two samples have the same mean values. We computed the statistical significance tests for TSCNN$_1$ and TSCNN$_2$ in each mode. In MI mode, the results showed a significant difference ($p=2.94\times10^{-4}$), while in SSVEP mode and hybrid mode, the results did not perform a significant difference ($p=0.048$ and $p=0.561$, respectively). This shows that TSCNN$_2$ retains most of the performance on SSVEP and Hybrid mode while significantly improving the decoding accuracy on MI. In addition, we also calculated the significant difference between TSCNN$_2$ and SCNN. Neither of them showed significant differences in MI nor SSVEP ($p=0.598$ and $p=0.008$, respectively). This proves that the decoding performance of TSCNN$_2$ in MI mode and SSVEP mode is similar to that of SCNN, which illustrates that TSCNN can maintain most of the performance in MI and SSVEP modes while having high decoding accuracy in hybrid mode.

For parameter choice, we also calculated the statistical significance of different dropout rates and activations.
Different dropout rates did not show significant differences in SSVEP and hybrid modes. There were significant differences between dropout of 50\% and 25\% ($p=0.004$) and dropout rate of 50\% and 0\% ($p=0.001$) in MI mode. 
Therefore, dropout rate of 50\% is adopted in TSCNN$_2$.
The choice of activation is also critical to the performance of the model~\cite{act2, act1}.
Different from dropout rate, none of the four different activations (ReLU, ELU, Softplus, LeakyReLU) show significant differences in the three modes ($p>0.2$). 

\subsection{Interpretation of Connection Weights}
The results clearly indicate that TSCNN$_2$ has better performance than TSCNN$_1$ in MI mode.
We hypothesize that the features of SSVEP are more obvious and cover up some MI features, resulting in a more prominent representation of SSVEP after fusion of two modes in TSCNN$_1$.
When trained with MI mode, the layers in the SSVEP block will not be activated, so the model will only fine-tune the kernel weights based on the results of the MI, thereby improving the representation of MI.

To verify our hypothesis, we need to compare the differences between the representations of MI and SSVEP in TSCNN$_1$ and TSCNN$_2$.
The concatenation layer connected to the fully connected layer has two feature maps, one of which is the output of the MI block and the other is the output of the SSVEP block.
We extracted the weights of the fully connected layer learned by TSCNN$_1$ and TSCNN$_2$, and use the weights from each feature map to the fully connected layer to quantify the corresponding representation. We set some thresholds and counted the number of connection weights that are higher than the specified threshold.

\begin{table}[htbp]
\centering
\caption{The number of weights that are greater than the threshold}
\begin{tabular}{ccccccc}
\hline
Model & Threshold & 0.0025 & 0.005 & 0.0075 & 0.01 & 0.0125\\
\hline
& $N_M$ & 29713 & 9699 & 2374 & 407 & 64\\
TSCNN$_1$ & $N_S$ & 30870 & 10674 & 2747 & 569 & 96\\
& $Ratio$ & 0.963 & 0.909 & 0.864 & 0.715 & 0.667\\
\hline
& $N_M$ & 30121 & 10219 & 2665 & 534 & 82\\
TSCNN$_2$ & $N_S$ & 30510 & 10089 & 2375 & 386 & 57\\
& $Ratio$ & 0.987 & 1.013 & 1.122 & 1.383 & 1.439\\
\hline
\end{tabular}
\label{thr}
\end{table}

\begin{figure}[ht]
    \centering
    \includegraphics[width=0.85\linewidth]{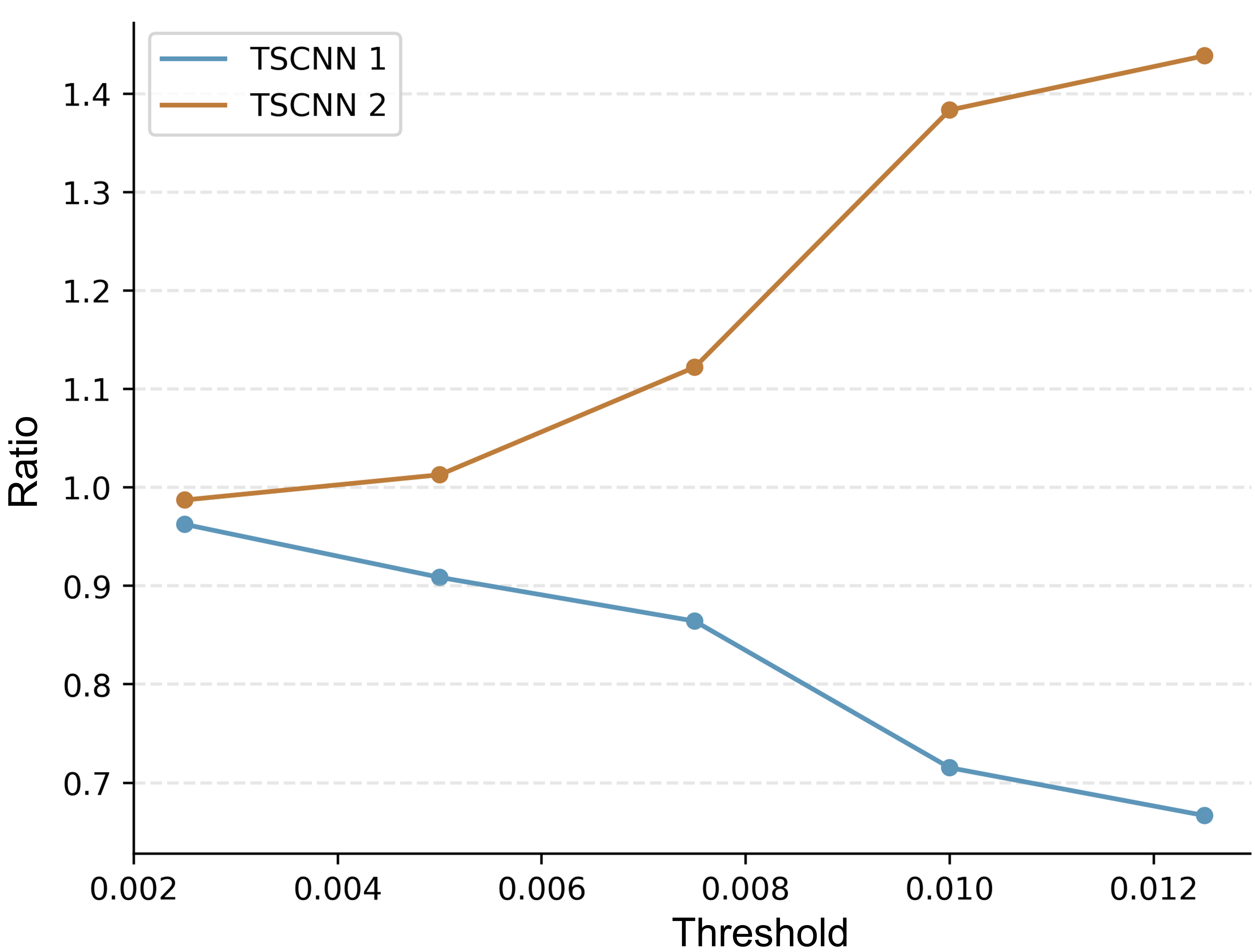}
    \caption{The ratio according to different thresholds. The thresholds are the weights of connections that are relatively high. The vertical axis is the ratio of the number of connections exceeding the threshold in TSCNN$_1$ to the number in TSCNN$_2$. The blue curve is the ratio in TSCNN$_1$, and the red curve is the ratio in TSCNN$_2$.}
    \label{con}
\end{figure}

The number of connections that are higher than the threshold of MI is denoted as $N_M$ and for SSVEP it is $N_S$. We compute the ratio of $N_M$ and $N_S$ to reflect the representation ability of the concatenation layer for MI.
$$ratio = \frac{N_M}{N_S}$$
The larger the ratio, the stronger the representation ability of the model for MI. Fig. \ref{con} illustrates the ratio change according to the increased threshold. As the threshold increases, the ratio of TSCNN$_1$ decreases while that of TSCNN$_2$ increases.

\section{Discussion}
In this study, we proposed a deep learning model called TSCNN$_2$, which can provide considerable decoding performance in both single-mode and hybrid-mode scenarios.

\subsection{From Methodology Perspective}
Universality and high-precision decoding performance are crucial for BCI systems.
However, previous hybrid BCI methods, such as linear discriminant analysis (LDA), CCA, and CSP, have unsatisfactory decoding accuracy and universality\cite{ko, Allison_2010}. As a deep learning model, TSCNN can achieve favorable performance in both single-mode and hybrid-mode scenarios, serving as a universal hybrid BCI framework.

Theoretically, TSCNN can be thought of as a combination of feature extraction and classification of the input. Compared with previous methods, the nonlinearity of TSCNN contributes to learning more complex latent representations and therefore higher performance of decoding. Moreover, TSCNN is well suited to both single-mode and hybrid-mode scenarios, while previous methods can only work in single-mode scenarios.

Another advantage of TSCNN is its interpretability, compared with other two-stream CNNs. We analyzed the connection weights in Section III-C. Table~\ref{thr} and Fig.~\ref{con} indicate that the fusion block has different representation abilities for the features of each stream,
and the differences can be adjusted by applying designed the training strategy, allowing the model to learn the feature of each stream more evenly. The two-stream architecture and the proposed training strategy together achieve satisfactory decoding accuracy and universality for hybrid BCI.

\subsection{From Application Perspective}
Compared to previous hybrid BCI methods,
TSCNN can achieve satisfactory decoding performance in both single-mode and hybrid-mode scenarios. This flexibility makes TSCNN a promising choice for use in hybrid BCI systems.
Subjects with damage to their motor or visual cortex may not be able to generate MI or SSVEP activity, respectively. However, these subjects can still use BCIs by utilizing the available mode. For instance, a subject with motor cortex damage can use the SSVEP mode to control the BCI, while a subject with visual cortex damage can use the MI mode. Additionally, subjects who are proficient in using both modes can generate MI and SSVEP activity simultaneously, allowing for high decoding accuracy. This demonstrates the versatility and adaptability of BCIs for a range of subjects and brain activity patterns.
Training and testing TSCNN on a variety of subjects allows for subject-independence, meaning that it is able to overcome inter-subject variability in EEG signals. This makes TSCNN a useful tool for addressing the challenges posed by individual differences in brain activity. Its ability to adapt to a range of subjects demonstrates its potential for use in a variety of contexts.

\subsection{Limitations and Future Works}
There are still some limitations to our work.
First, 
The ability to achieve high decoding accuracy of EEG signals with short epoch lengths is essential for the development of effective BCI systems. A shorter epoch length allows for higher temporal resolution, enabling the detection of changes in brain activity at a faster rate. However, in this study, our proposed TSCNN method was trained and tested with 4-second EEG. Further research is needed to investigate the potential of methods that can achieve high decoding accuracy with even shorter epoch lengths.
Second, TSCNN is a fully supervised deep learning model, and thus, its training process requires a large amount of labeled data. Supervised deep learning models are highly dependent on the quality and relevance of the training data, and may not generalize well to unseen data. In order to further verify the effectiveness of our proposed method, it is important to conduct additional experiments using expert-annotated datasets. In the future, we also plan to explore unsupervised, semi-supervised, and domain generalization methods as potential avenues for improving the performance of our model.
Finally, in our proposed TSCNN model, the MI block and SSVEP block used the same architecture, allowing them to learn the features of their respective modalities. However, there are already many deep learning models that have demonstrated the performance in single-mode MI and SSVEP decoding~\cite{subj,Sch, Chiang_2021}. Therefore, further research is needed to investigate fusion frameworks that can achieve higher performance by using these high-performing models architecture in a fusion context.

\section{Conclusion}
We proposed a TSCNN framework using SSVEP and MI for a hybrid BCI system to achieve high versatility and generalization. TSCNN can automatically extract EEG features in both paradigms. We presented a new training strategy for hybrid BCI models, which achieve superior performance in both uni-mode and hybrid-mode scenarios.

\ifCLASSOPTIONcaptionsoff
  \newpage
\fi

\bibliography{ref}

\end{document}